\documentclass[letterpaper]{article} 
\usepackage{aaai24}  
\usepackage{times}  
\usepackage{helvet}  
\usepackage{courier}  
\usepackage[hyphens]{url}  
\usepackage{graphicx} 
\usepackage{natbib}  
\usepackage{caption} 
\usepackage{amsmath}
\usepackage{algorithm}
\usepackage{algorithmicx}
\usepackage{multirow}
\usepackage{adjustbox}
\usepackage{amssymb}

\usepackage{newfloat}
\usepackage{listings}
\DeclareCaptionStyle{ruled}{labelfont=normalfont,labelsep=colon,strut=off} 
\floatstyle{ruled}
\newfloat{listing}{tb}{lst}{}
\floatname{listing}{Listing}
\title{
BrainNetDiff: Generative AI Empowers Brain Network Generation \\ via Multimodal Diffusion Model
}
\author{
    Yongcheng Zong\textsuperscript{\rm 12}, Shuqiang Wang\textsuperscript{\rm 1}\thanks{Corresponding Author.},
}
\affiliations{

    \textsuperscript{\rm 1} Shenzhen Institutes of Advanced Technology, Chinese Academy of Sciences  \\
    \textsuperscript{\rm 2} University of Chinese Academy of Sciences \\
    yc.zong@siat.ac.cn \\
    sq.wang@siat.ac.cn
}

\usepackage{bibentry}

\begin{document}

\maketitle

\begin{abstract}
Brain network analysis has emerged as pivotal method for gaining a deeper understanding of brain functions and disease mechanisms. Despite the existence of various network construction approaches, shortcomings persist in the learning of correlations between structural and functional brain imaging data. In light of this, we introduce a novel method called BrainNetDiff, which combines a multi-head Transformer encoder to extract relevant features from fMRI time series and integrates a conditional latent diffusion model for brain network generation. Leveraging a conditional prompt and a fusion attention mechanism, this method significantly improves the accuracy and stability of brain network generation. To the best of our knowledge, this represents the first framework that employs diffusion for the fusion of the multimodal brain imaging and brain network generation from images to graphs.
We validate applicability of this framework in the construction of brain network across healthy and neurologically impaired cohorts using the authentic dataset. Experimental results vividly demonstrate the significant effectiveness of the proposed method across the downstream disease classification tasks. These findings convincingly emphasize the prospective value in the field of brain network research, particularly its key significance in neuroimaging analysis and disease diagnosis. This research provides a valuable reference for the processing of multimodal brain imaging data and introduces a novel, efficient solution to the field of neuroimaging.

\end{abstract}

\section{Introduction}
In recent years, the rapid advancements in neuroimaging and artificial intelligence have presented new avenues for comprehensively understanding brain function and the underlying mechanisms of neurological disorders, including Alzheimer's disease \cite{minati2009reviews, zuo2021prior}, Parkinson's disease \cite{Parkinson}, and autism, which profoundly impact individuals' quality of life and well-being. To further our understanding, diagnosis, treatment, and exploration of these neurological disorders, medical imaging analysis techniques have become indispensable tools. Particularly, methods grounded in artificial intelligence \cite{wang2012bayesian}, notably utilizing deep learning models, have made remarkable strides in the realm of medical image processing, offering substantial potential for medical diagnosis and research.

The construction and analysis of brain networks have emerged as pivotal methodologies. Among various neuroimaging modalities, diffusion tensor imaging (DTI) and functional magnetic resonance imaging (fMRI) have been widely employed for the construction of brain networks \cite{wang2020ensemble}. In these modalities, nodes are typically defined as regions of interest (ROIs) on a given brain atlas, while edges are computed based on the number of neural fiber connections for structural brain networks or pairwise correlations of blood oxygen level-dependent (BOLD) signal sequences extracted from each region for functional brain networks \cite{2022Predicting}. However, challenges persist in the fusion of multimodal brain imaging data and the generation of brain networks, entailing intricate complexities.

At present, diffusion-based models \cite{croitoru2022diffusion, rombach2022high} have demonstrated substantial success in various downstream tasks within domains such as natural language processing and computer vision. This study is inherently motivated by this prevailing context. In response, we present an innovative approach that amalgamates a multi-head Transformer Encoder and a conditional latent diffusion model to enhance the precision and stability of brain network generation. By harnessing a conditional prompt and a fusion attention mechanism, BrainNetDiff effectively integrates functional time series features with brain network structures, thereby introducing a novel pathway for the construction of brain networks.

The main contributions of this paper can be summarized as follows:

\begin{itemize}
\item We introduce a conditional latent diffusion model, which offers a novel paradigm for fusing structural-functional imaging and generating brain networks. Compared with traditional voxel-based methods, BrainNetDiff significantly improves computational efficiency while maintaining the quality of the generated brain networks.

\item Furthermore, by employing fMRI time series embedding as a conditional prompt and incorporating attention mechanisms, we successfully fuse structural and functional brain images in latent space, providing robust support for generating high-quality brain networks.

\item Finally, we use contrastive joint pretraining to establish correlations between images and brain networks. By exploiting the similarity features, we conduct the self-supervised training for image tasks. The classification task is transformed into an image-graph matching task, which outperforms the separate self-supervised training methods.
\end{itemize}

We extensively validate the effectiveness and superiority of our proposed BrainNetDiff model through experiments on real dataset. The results demonstrate our method's exceptional performance in generating high-quality brain networks. Compared to existing approaches, BrainNetDiff exhibits higher efficiency and superior fusion performance, underscoring its crucial application value in processing multimodal brain imaging data.

\begin{figure*}[t]
\centering
\includegraphics[width=\textwidth]{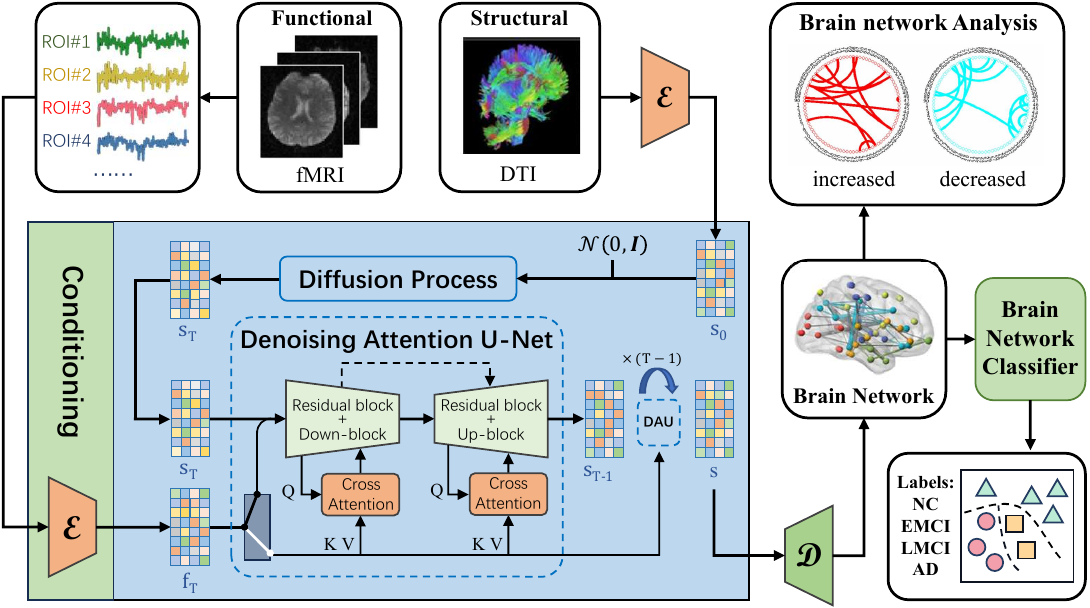} 
\caption{The network architecture of the proposed BrainNetDiff has three sub-modules: (a)Image encoder and graph encoder, (b)brain feature learning in the latent space and (c)brain network generation.}
\label{fig1}
\end{figure*}

\section{Related Works}
\subsection{Brain Network Analysis}

Recently, there has been growing attention towards the application of GNN-based models in brain network analysis \cite{ahmedt2021graph} . GroupINN \cite{groupinn} employs grouped layers to enhance interpretability and reduce model size. BrainGNN \cite{braingnn} devises a GNN that captures functional information of Regions of Interest (ROIs) in brain networks, employing specialized pooling operators to select pivotal nodes. \cite{cui2022} introduces an interpretable framework for analyzing disease-specific ROIs and significant connections. Additionally, \cite{fbnetgen} delves into learnable brain network generation, investigating the interpretability of generated networks for downstream tasks. Until recent years, similar work \cite{braingb} systematically explores diverse GNN designs for brain network data. Unlike efforts focusing on static brain networks, STAGIN \cite{kim2021learning} models dynamic brain networks extracted from fMRI data using GNNs empowered with spatiotemporal attention. In summary, the above studies have made significant contributions to the field of brain network construction and analysis, particularly for Alzheimer's disease diagnosis and treatment. However, these methods still have limitations, such as the dependence on templates, lack of interpretability, and sensitivity to noise.

\subsection{Multimodal Fusion and Learning}

In the realm of disease prediction, early multimodal studies \cite{pham2007, lee2009} typically fuse raw multimodal features via direct concatenation. However, a challenge arises from the heterogeneous nature of such data, potentially leading to the curse of dimensionality. To address this, methods like \cite{zhang2011, liu2013} apply multi-kernel learning to capture kernels for each modality, which are then linearly or weight-wise combined. Furthermore, \cite{kang2020diagnosis, pan2020multi} encode raw features of each modality via DNNs and concatenate embeddings to obtain fused representations. However, these methods often neglect modal interactions, lacking an effective exploration of inter-modality correlations. To this end, \cite{zhou2019} leverages deep semi-non-negative matrix factorization to learn shared modality representations. To delve deeper into modality complementarity, \cite{ning2021} establishes a bidirectional mapping between raw features and shared embeddings to preserve original information. However, the effective exploration of inter-modality information in multimodal data remains a pivotal challenge for disease diagnosis. Beyond multimodal interplay, patient relationships should also be considered.

Given graphs' potent expressive capabilities in modeling relationships, graph-based approaches have gained traction in disease prediction. \cite{tong2017} constructs graphs for each modality using handcrafted kernels, subsequently merging them for classification. \cite{gao2020} employs GCNs to learn embeddings of patients on constructed multi-graphs. In \cite{parisot2017, kazi2019}, non-imaging features construct a global graph, while imaging features act as individual features for neighbor aggregation. Additionally, \cite{valenchon2019, kazi2019self} introduce attention mechanisms for multimodal fusion in \cite{tong2017}. Notably, these methods often separate handcrafted graph construction from prediction modules, leading to cumbersome adjustments and subpar generalization. Moreover, they inadequately explore inter-modality relationships. Inspired by these observations, BrainNetDiff is proposed to capture shared modality information and modality-specific information synchronously within an end-to-end adaptive graph learning framework.

\subsection{Graph Generation and Diffusion Generation}
Graph generation models generate all edges between nodes at once. VAEs \cite{dai2019} and GANs \cite{generative} models generate all edges independently of latent embeddings. However, this independence assumption may compromise the quality of generated graphs. Normalizing flow models \cite{zang2020} are limited to reversible model architectures for constructing normalized probabilities. Another category is autoregressive graph generation models, generating graphs by sequentially adding nodes and edges. Autoregressive generation can be achieved using recursive networks \cite{li2018learning}, VAEs \cite{liu2018constrained}, normalizing flows \cite{shi2001graphaf}, and reinforcement learning \cite{you2018graph}. These methods, by breaking down the problem into smaller parts, are more adept at capturing complex structural patterns, easily incorporating constraints during the generation process.  Existing research \cite{lucas2019understanding} indicates that VAE objectives can lead to posterior collapse.

Diffusion models have emerged as powerful deep generative models. Denoising Diffusion Probabilistic Models (DDPM) \cite{ho2020denoising} perturb data distribution into a Gaussian distribution via forward Markov noise, subsequently learning to recover data distribution through reverse transitions in the Markov chain. Closely related to DDPM is score-based generation \cite{song2020score}, which perturbs data by gradually increasing noise and learns inverse perturbations through score matching. Song et al. (2021) extend diffusion models to continuous-time diffusion using forward and backward SDEs. Existing diffusion-based graph generation models have been one-shot. Niu et al. employ score matching with varying noise scales to model the adjacency matrix \cite{niu2020permutation}, generating using annealed Langevin dynamics. In the subsequent work, DiGress \cite{vignac2022digress} further considered generating node feature vectors and graph topology together. The existing diffusion based graph generation models are one-time. Inspired by this, the model proposed in this article uses fMRI data as conditional diffusion to jointly model the adjacency matrix and node features, thereby obtaining higher quality and more stable brain networks.

\begin{figure}[t]
\centering
\includegraphics[width=0.5 \textwidth]{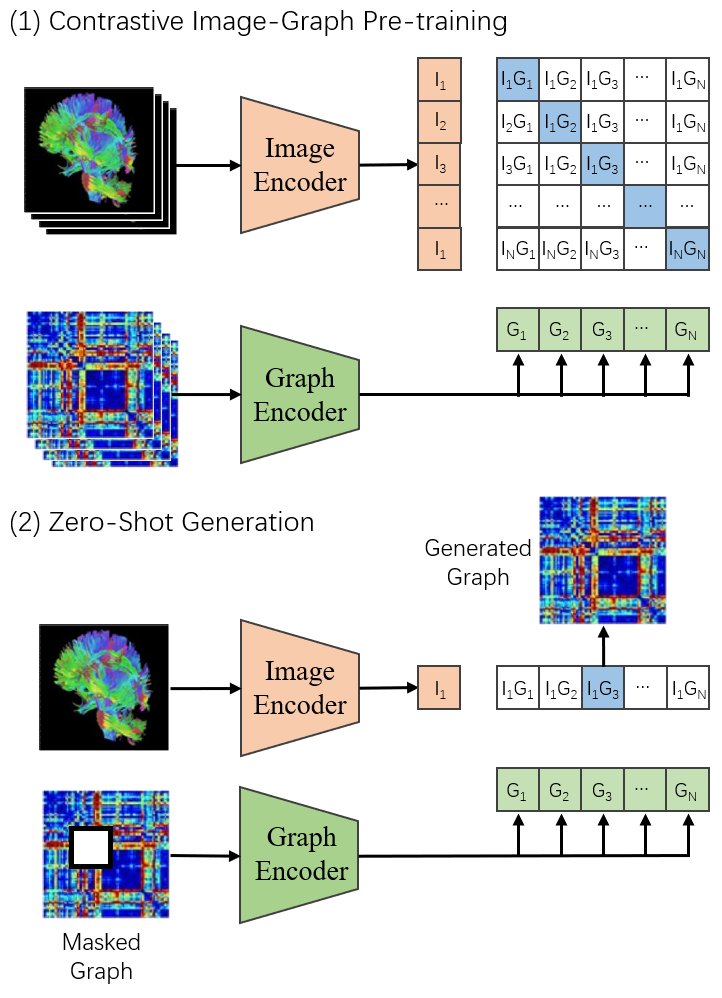} 
\caption{CIGP uses graphs as the supervised signal to train a transferable Image encoding model. Inspired by the CLIP method, it achieves image to graph matching and is used for zero-shot brain network generation.}
\label{fig2}
\end{figure}

\section{Methods}
\subsection{Conditional Latent Diffusion Model}

Figure \ref{fig1} presents the overall model architecture.
The Diffusion Model (DM) is a probabilistic model based on Markov chains that learns the data distribution $p(x)$ by iteratively denoising a normally distributed variable. Widely applied in image synthesis and generation tasks, these models act as denoising autoencoders, $\epsilon_\theta(x_t, t)$, with the corresponding objective:
\begin{equation}
  L_{D M}=E_{x, \epsilon \sim \mathcal{N}(0,1), t}\left\|\epsilon-\epsilon_\theta\left(x_t, t\right)\right\|_2^2
\end{equation}
The Latent Diffusion Model (LDM) denoises images in the latent space encoded by the Encoder, accessing a lower-dimensional latent space. Compared to direct manipulation in pixel space, lower-dimensional spaces offer greater computational efficiency and better suit likelihood-based generative models that focus on essential semantic features.
\begin{equation}
    L_{L D M}=E_{\mathcal{E}(x), \epsilon \sim \mathcal{N}(0,1), t}\left\|\epsilon-\epsilon_\theta\left(z_t, t\right)\right\|_2^2
\end{equation}

The proposed BrainNetDiff module interconnects multiple Transformer-based cross-attention mechanisms to modulate LDM. It chiefly leverages the capacity of 2D convolutional layers to construct the underlying UNet and employs reweighted boundaries to further focus on regions of crucial relevance. The forward process involves latent $z_t$ obtained from the Encoder and samples of $p(z)$, which are subsequently decoded to brain network space by the Decoder.

Diffusion models can simulate conditional distributions of the form $p(z|y)$. Innovatively, we utilize each sample's corresponding fMRI data as guiding conditions. This facilitates stable brain network generation by employing the conditional denoising autoencoder $\epsilon_t(z_t, t, y)$ to fuse additional disease information.
Inspired by related literature \cite{radford2021learning}, we enhance the Diffusion Network's UNet backbone using cross-attention mechanisms. This transforms LDM into a more flexible conditional image generator, accommodating various input modes and extending to other conditional guidance. To address time-series fMRI data, a domain-specific encoder is introduced, projecting condition y into an intermediate representation $\theta(y)$, which is then mapped to the UNet's intermediate layers through cross-attention layers, facilitating multimodal fusion with the attention mechanism.
\begin{equation}
    \operatorname{Attention}(Q, K, V)=\operatorname{softmax}\left(\frac{Q K^T}{\sqrt{d}}\right) \cdot V
\end{equation}
where $Q=W_Q^{(i)}\cdot\varphi_i(z_t),K=W_K^{(i)}\cdot\theta(y),V=W_V^{(i)}\cdot\theta(y)$.

Here $\varphi_i(z_t)$ denotes the UNet's intermediate layer representation, and $W$ represents learnable projection matrices. Given the conditional mechanism, our objective function becomes:
\begin{equation}
    L_{LDM}=\mathbb{E}_{\mathcal{E}(x), y,\epsilon\sim\mathcal{N}(0,1),t}\left[{\parallel\epsilon-\epsilon_\theta(z_t, t, \theta(y))\parallel}_2^2\right]
\end{equation}

\subsection{Contrastive Image-Brain Network Pretraining}
To conduct diffusion learning in the latent space, we must encode images and brain networks. Unlike natural images with three channels, used DTI data is completely three-dimensional data. Considering computational complexity and training time, a suitable network structure must be designed to process three-dimensional data. Recent research in image contrastive representation learning suggests that using a contrastive objective function constraint is superior to its equivalent predictive objective for learning representations. Moreover, compared to image generation models, equivalent-performance contrastive models require an order of magnitude less computation. Hence, we design a more straightforward surrogate task, predicting which image pairs correspond to a brain network as a whole. By means of contrastive joint pretraining of brain images and brain networks, the classification task is transformed into an Image-Graph matching task, effectively enhancing the model's performance in image self-training.

We employ ResNet-50 as the foundational architecture for image encoding, as it is widely adopted and mature in performance. We replace the global average pooling layer with an attention pooling mechanism, wherein the query conditions on the globally averaged pooled image representation. This enables the learning of global information in the network's shallow layers.

The graph (brain network) encoder is a 4-layer GAT autoencoder, enhancing node representations through multiple attention heads. For node i, the similarity coefficient between its neighbors ($j \in N_i$) and itself is calculated individually:
\begin{equation}
    e_{ij}=a([Wh_i\parallel Wh_j]),j\in N_i
\end{equation}

The above linear mapping enhances vertex features by dimensionality augmentation. The attention coefficient is defined as:

\begin{equation}
    \alpha_{i j}=\frac{\exp \left(\operatorname{LeakyReLU}\left(e_{i j}\right)\right)}{\sum_{k \in \mathcal{N}_i} \exp \left(\operatorname{LeakyReLU}\left(e_{i k}\right)\right)}
\end{equation}

Node propagation in GAT uses multi-head attention for aggregation. In contrast to GCN, it employs vertex-wise operations for weighted summation. By calculating attention over adjacent nodes, the correlation between vertex features is better integrated into the model. The graph's embedding representation $z_g$ is obtained via Readout.
\begin{equation}
    h_i^\prime(K)=\parallel_{k=1}^K\sigma\left(\sum_{j\in\mathcal{N}_i} \alpha_{ij}^kW^kh_j\right)
\end{equation}

Figure \ref{fig2} depicts the encoder's pretraining schematic.

\subsection{Conditional Guidance}

Most generative models (such as GANs and flow-based models) often possess the ability to perform truncated sampling by reducing the variance or range of noise inputs during generation, thereby decreasing sample diversity while enhancing the quality of each generated sample. In the Diffusion model, direct truncation operations cannot be employed. In this study, we employ a classifier-guided approach to enhance stability in the latent space.

The gradient of the log-likelihood of the auxiliary classifier model $p_\theta\left(\mathbf{c}\mid\mathbf{z}\lambda\right)$ is used to approximate the diffusion score:

\begin{equation}
\begin{aligned}
\widetilde{\epsilon}_\theta\left(\mathbf{z}_\lambda, \mathbf{c}\right) & =\epsilon_\theta\left(\mathbf{z}_\lambda, \mathbf{c}\right)-w \sigma_\lambda \nabla_{\mathbf{z}_\lambda} \log p_\theta\left(\mathbf{c} \mid \mathbf{z}_\lambda\right) \\
& \approx-\sigma_\lambda \nabla_{\mathbf{z}_\lambda}\left[\log p\left(\mathbf{z}_\lambda \mid \mathbf{c}\right)+w \log p_\theta\left(\mathbf{c} \mid \mathbf{z}_\lambda\right)\right]
\end{aligned}
\end{equation}
Here, $w$ is a parameter controlling the classifier-guided strength. This modified score is sampled from the diffusion model, yielding approximate samples from the distribution.

Therefore, the objective function of our conditional diffusion model is
\begin{equation}
    L_{reconstruction}=\mathbb{E}_{\mathcal{E}(x), y,\epsilon\sim\mathcal{N}(0,1),t}\left[{\widetilde{\epsilon}_\theta\left(\mathbf{z}_\lambda, \mathbf{c}\right) -\epsilon\parallel}_2^2\right]
\end{equation}

Classifier guidance enhances the probability of correct label assignment for data, leading to stronger influence of higher-confidence guidance information on generated brain networks. This improves the stability and accuracy of brain network generation. The training process is as follows:

\begin{algorithm}[!h]
\caption{Joint training a diffusion model with classifier-free guidance}
\begin{algorithmic}[1]
\item $ \text { Require: } p_{\text {cond }} \text { probability of conditional training }$ \\
$\text { Repeat }  $\\
\quad $(\mathbf{x}, \mathbf{c}) \sim p(\mathbf{x}, \mathbf{c}) $ \\
\quad $ \mathbf{c} \leftarrow \varnothing \text { with probability } p_{\text {cond }}$  \\
\quad $ \lambda \sim p(\lambda) $ \\
\quad $ \boldsymbol{\epsilon} \sim \mathcal{N}(\mathbf{0}, \mathbf{I}) $ \\
\quad $ \mathbf{z}_\lambda=\alpha_\lambda \mathbf{X}+\sigma_\lambda \boldsymbol{\epsilon} $ \\
$\text { Take gradient step on } \nabla_\theta\left\|\epsilon_\theta\left(\mathbf{z}_\lambda, \mathbf{c}\right)-\boldsymbol{\epsilon}\right\|^2 \text {  until converged }$ 
\end{algorithmic}
\end{algorithm}

\subsection{Classifier Design and Loss Function}
We design a classifier for multi classification of generated brain networks. The classifier aggregates through multi-layer perceptrons, containing graphical representations of rich pathological information, and finally connects with the Softmax function to output the prediction probability of each category of diseases. This module consists of a three-layer backpropagation neural network, two layers of ReLu activation function and Softmax output layer. The Softmax input contains four neurons, mapping the results to the probabilities of four disease categories. The Dropout strategy is used to prevent the model from overfitting \cite{dropout}.

Multi-class cross entropy function is adopted for multi classification loss \cite{lian2019end}. In our study, the number of classes $C$ is 4.
\begin{equation}
\mathcal{L}_{\mathrm{classification}}=-\sum_{i=0}^{C-1}{y_i\log (p_i)}+({1-y}_i)\log (1-p_i)
\end{equation}

At last, the overall loss function of the proposed model is as follows:
\begin{equation}
{\mathcal{L}_{total}=\mathcal{L}_{\mathrm{reconstruction}}+\mathcal{L}}_{\mathrm{classification}}
\end{equation}

\bigskip
\section{Experiments}
\subsection{Dataset and Preprocessing}
This study utilize diffusion weighted imaging (DWI) and T1-weighted MRI data from the Alzheimer's Disease Neuroimaging Initiative (ADNI) \cite{2010Alzheimer}, an open-source and public dataset to validate the proposed framework . A total of 349 subjects' data were collected, including normal control group (NC), early mild cognitive impairment (EMCI), late mild cognitive impairment (LMCI), and AD. Table 1 provides detailed information about the sample size, gender, and age of all subjects. The PANDA toolbox \cite{2013PANDA} was used for preprocessing the raw DTI data to obtain the reference structural brain network matrix. The process involved converting the initial DICOM format of the data to NIFTI format, skull stripping, fiber bundle resampling, and head motion correction. Then we calculate the fractional anisotropy (FA) coefficients by fitting the tensor model using the least squares method, and output the DTI data. After resampling, all subjects had $91 \times 109 \times 91$ voxels in DTI, with voxel size of $2mm \times 2mm \times 2mm$. After that, we registered T1 image to the individual brain space, and constructed the empirical structural connectivity $\hat{A}$ based on the deterministic fiber tracking method by setting tracking conditions, network nodes, and tracking stopping conditions. The brain was divided into 90 regions of interest (ROIs) based on the AAL atlas \cite{2019Automated}, with each ROI defined as a node in the brain network. Finally, the structural connectivity of the brain network was determined by fiber tracking between different ROIs. Specifically, the stopping criteria for fiber tracking were defined as follows: (1) the crossing angle between two consecutive directions is greater than 45 degrees, and (2) the anisotropy score value is not within the range of [0.2, 1.0].

\begin{table}[!ht]
\adjustbox{width=\columnwidth}{
    \centering
    \begin{tabular}{ccccc}
    \hline
        Group & NC(87) & EMCI(135) & LMCI(63) & AD(64) \\ 
        \hline
        Gender & 46M/41F & 83M/72F & 26M/37F & 35M/29F \\ 
        Age & 74.3$\pm$5.5 & 74.9$\pm$5.8 & 75.8$\pm$6.1 & 75.6$\pm$5.4 \\
        \hline
    \end{tabular}
}
\caption{Subjects’ information in this study.\label{table1}}
\end{table}

\subsection{Experiment Settings}
The model was trained and tested using the PyTorch platform, with an NVIDIA RTX 4000 GPU with 20GB memory. During the training process, the optimizer was set to Adam \cite{2014Adam}, with an initial learning rate of 0.0001, which exponentially decayed with the number of training iterations. The number of epochs was set to 300, and the batch size was set to 16. Five-fold cross-validation was used to evaluate the performance of the model. All subjects were randomly divided into five equally sized subsets. One subset was treated as the test set, and the union of the other four subsets was treated as the training set. This process was repeated five times to eliminate bias. The evaluation metrics for the classification performance in this study were accuracy (ACC), sensitivity (SEN), specificity (SPE), and the area under the ROC curve (AUC).

The learnable parameter is initialized according to the Xavier scheme. The structure of BrainNetDiff has been described in detail in section Methods. We use two Encoders to progressively extract high-order topological features of brain regions, with a brain region feature vector dimension of $d=128$. The learnable parameter is initialized with the identity matrix. The Discriminator consists of two FC layers, with a LeakyReLU activation function in the hidden layers and no activation function in the output layer. The classification feature $h \in \mathbb{R}^{90 \times 128}$ is inputted into the FC layer for disease prediction, with 128 neurons in the input layer and 4 neurons in the output layer.

In section of \textbf{Hyperparameter select}, we analyze the impact of hyperparameters and performance on the classification results. In section of \textbf{Classification performance}, we compare the classification performance of the structural brain networks generated by our proposed model with those constructed by PANDA software using GCN models. Section of \textbf{Brain structural network connectivity analysis} demonstrates the superiority of our model over software templates in terms of brain network difference. Then we focus on the quantitative analysis of the brain networks generated by our model, aiming to explore the highly correlated abnormal connections and brain regions with the disease.

\bigskip
\subsection{Results and Discussion}
\subsubsection{Hyperparameter selection}
During the training process, the number of steps in the denoising process $T=300$, $\beta_t=\frac{0.02-0.0001}{T} t$.In this study, our main task is to use the subject's fMRI embedding as a conditional prompt to achieve deep fusion of structural functional brain networks, and to perform disease prediction and abnormal brain connectivity analysis. We use a stratified sampling method to select 30\% of the samples from all subjects as the test set. The remaining samples are used to train the model through 5-fold cross validation to search for the optimal value of hyperparameter embedding dimension $d$.

\begin{figure}[t]
\centering
\includegraphics[width= \columnwidth]{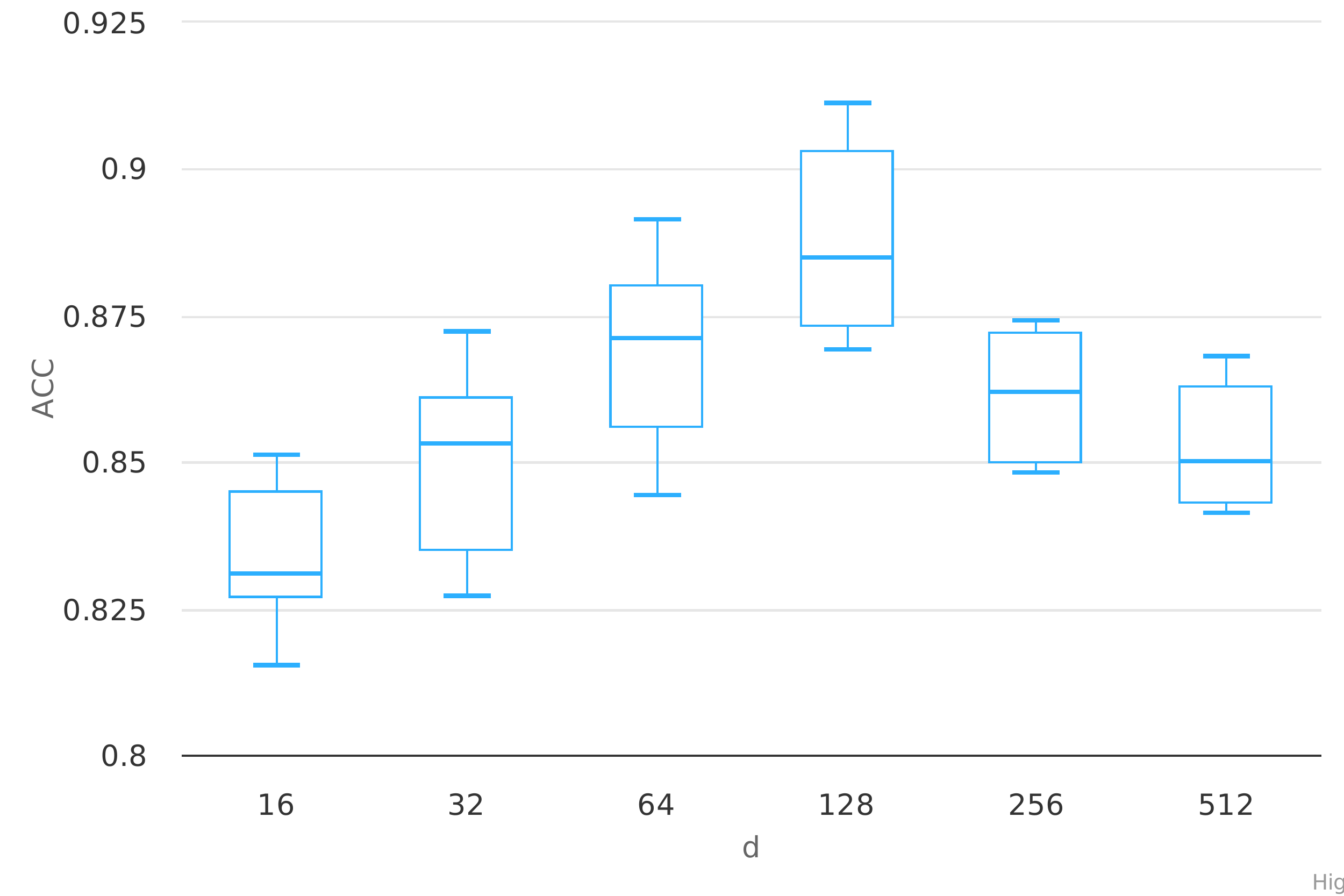} 
\caption{The impact of hyperparameters on model performance.}
\label{fig3}
\end{figure}


    

\subsubsection{Classification performance}
We compared BrainNetDiff with two other GNN based SOTA models as baselines. From the results in Table 2, it can be seen that the proposed BrainNetDiff performs significantly better than the GNN based GIN and GAT on the ADNI dataset, with an improvement of up to 11\% on the dataset. The input graphs processed by existing GNN models are usually not fully connected, and due to the design flaws of GNN in the nature of brain networks, it is difficult for GNN models to learn complex high-order network structures. Specifically, due to the fact that the brain network is a complete graph, key designs such as central encoding and spatial encoding of GNN cannot be applied appropriately. In addition, the preprocessing and training stages of the GNN model only accept discrete classification data, and then apply propagation operations on the adjacency matrix. Using the features extracted from structural images as coarse node features cannot learn the correlation and private features between structural and functional images.

\begin{table}[!ht]

\adjustbox{width=\columnwidth}{

    \centering
    \begin{tabular}{ccccc}
    \hline
        Method & ACC & AUC & SEN & SPE\\ 
    \hline
        GIN & 78.35$\pm$ 5.63 & 87.57$\pm$ 8.84 & 88.50$\pm$ 5.30 & 86.41$\pm$ 12.01 \\ 
        GAT & 76.48$\pm$ 5.16 & 83.26$\pm$ 5.23 & 85.07$\pm$ 6.44 & 87.19$\pm$ 13.31 \\ 
    \hline
        Ours & 86.70$\pm$ 3.68 & 92.22$\pm$ 8.16 & 91.86$\pm$ 4.29 & 89.77$\pm$ 10.48 \\ 
    \hline
    \end{tabular}
}
\caption{Quantitative evaluation on prediction performance(\%) of different methods, and values are reported as mean ± standard deviation. \label{tab:table6}}
\end{table}

\begin{figure}[t]
\centering
\includegraphics[width=0.5 \textwidth]{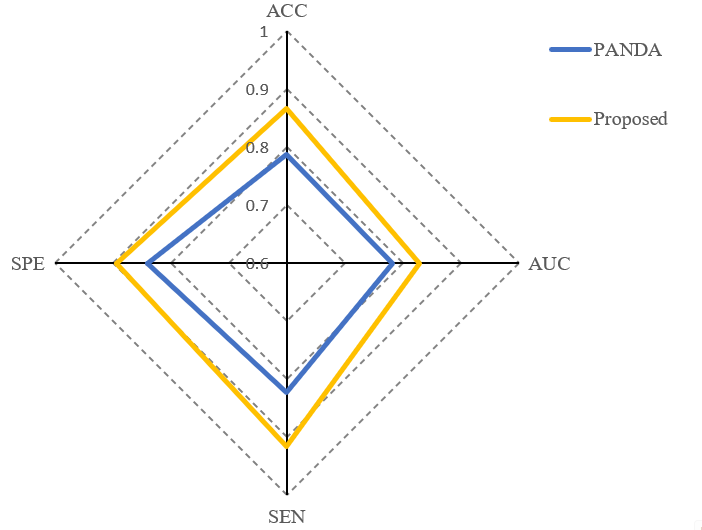} 
\caption{Radar charts of classification performance.}
\label{fig4}
\end{figure}

We also conducted comparative  experiments on the proposed Diffusion module. Specifically, we only use the brain network constructed by PANDA software as input to the classifier to compare classification performance. The experimental results are shown in the radar charts in Figure \ref{fig4}.

\subsubsection{Brain structural network connectivity analysis}
\begin{figure*}[ht]
\centering
\includegraphics[width=0.8 \textwidth]{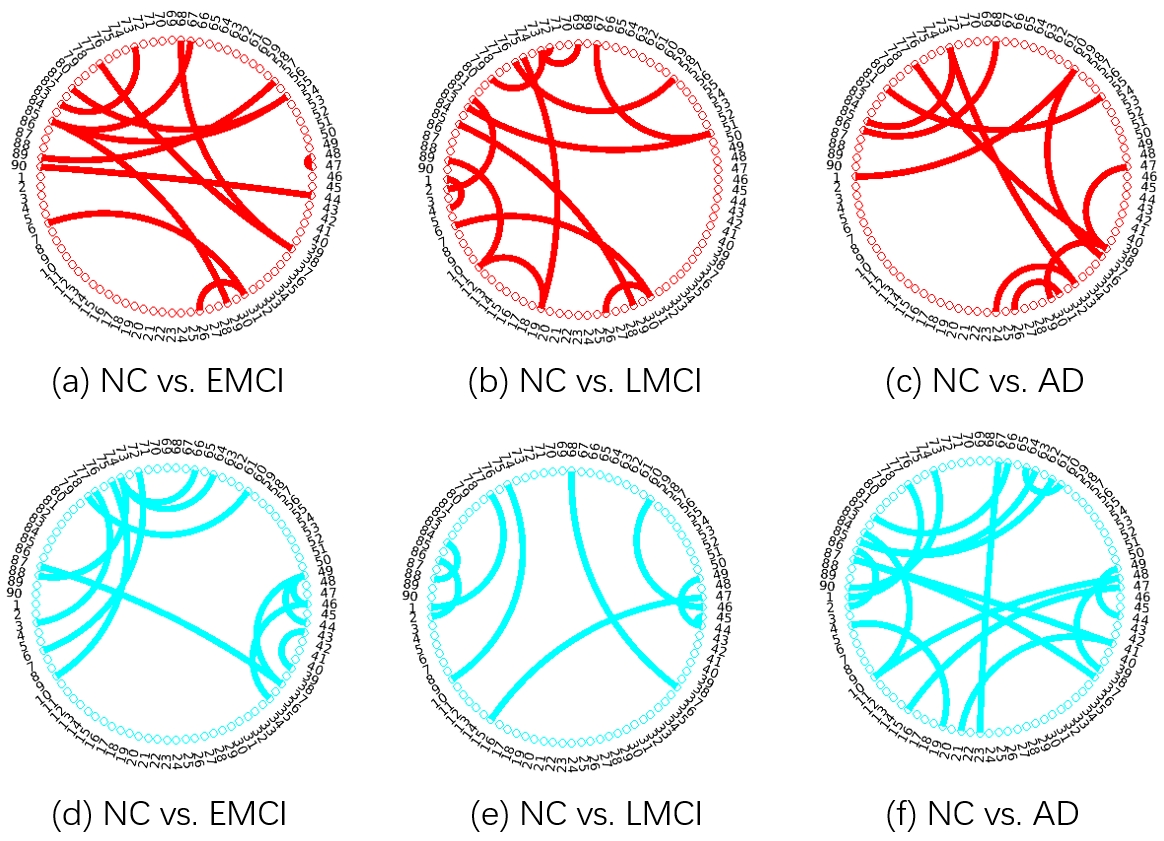} 
\caption{(a)-(c) represents the brain connections with increased connectivity, and (d)-(f) represents the brain connections with decreased connectivity.}
\label{fig5}
\end{figure*}

In order to determine whether the structural connectivity matrix $\hat A$ generated by our proposed model is significantly different from PANDA result  $A$, we conducted a dual sample T-test with a threshold of 0.05 between  $\hat A$ and  $A$ according to the settings in \cite{zong2022}. The chord diagram in Figure \ref{fig5} represents connections with significant changes. Compared to static software generation, the structural connection matrix  $\hat A$ derived from our model alters the connections between many brain regions.

\begin{figure}[t]
\centering
\includegraphics[width=\columnwidth]{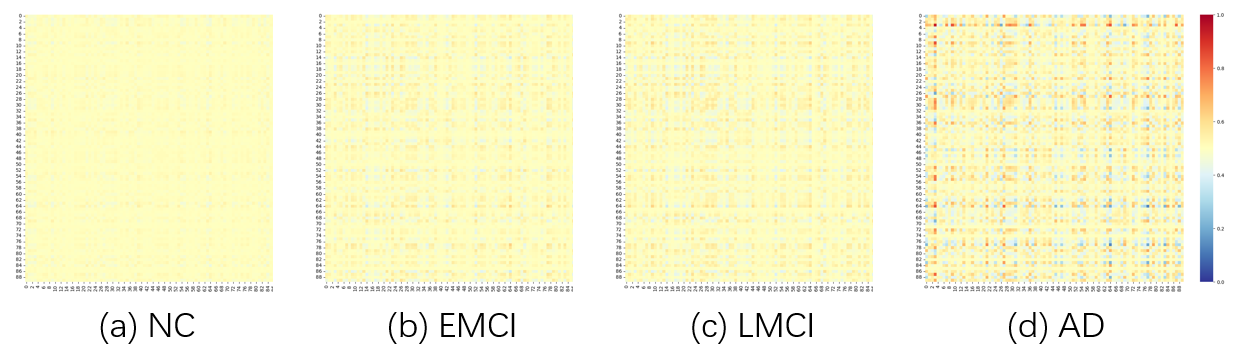} 
\caption{The difference of the constructed structural brain network at various developmental stages.}
\label{fig6}
\end{figure}

Specifically, the changes in connections between brain regions during the LMCI phase are more significant than those during the NC phase. Similarly, compared to EMCI patients, LMCI patients showed a significant decrease in connectivity between brain regions. These changes and trends reveal the continuous progression of NC subjects towards AD pathology: a gradual decrease in brain structural network connectivity,  which is consistent with existing literature on neuroscience research \cite{lei2022longitudinal}.

Figure \ref{fig6} shows the differences between the brain network generated by the proposed BrainNetDiff and the reference brain network provided by the template. From the figure, it can be seen that in the early stages of AD development, the learned brain network is very consistent with the connection matrix of the reference brain network, proving the effectiveness and consistency of our model. In the AD population, there is a significant deviation between our brain network and the reference template output, which is due to the fact that our connection matrix references more information about functional modalities, resulting in a significant difference in the final generated matrix.

\section{Conclusion}
This paper proposes the BrainNetDiff model, which combines a multi-head transformer encoder to extract relevant features from fMRI time series and integrates a conditional latent diffusion model for brain network generation. By utilizing conditional prompts and fused attention mechanisms, BrainNetDiff significantly improves the accuracy of brain network generation. We validated the performance of our model in constructing brain networks for healthy and neurocompromised populations on the real dataset, ADNI. The experimental results show that our method performs better in tasks of classifying different types of diseases. These findings emphasize the potential value of diffusion models in the field of brain network research, providing valuable references for the processing of multimodal imaging data, and introducing a new and effective solution for the field of neuroimaging. In future work, we will improve the brain network module and serve as the backbone for further brain network analysis, such as exploring the basic neural circuits and the developmental stages of cognitive disorders.


\end{document}